\documentclass[times, 10pt,twocolumn]{article}
\usepackage{latex12}
\usepackage{times}
\usepackage{balance}

\usepackage{amssymb}

\usepackage{color}
\usepackage{booktabs}
\usepackage{multirow}

\usepackage{epsfig} 
\usepackage{mathrsfs}
\usepackage{amsmath}
\usepackage{subfig}
\usepackage{graphicx}
\usepackage{enumerate}

\begin{document}

\title{Age Group and Gender Recognition From Human Facial Images}

\author{Tizita Nesibu Shewaye \\
\emph{Université Toulouse III - Paul Sabatier}\\
\emph{118 route de Narbonne 31062 TOULOUSE CEDEX 9}\\
\emph{tizita.nesibu@gmail.com}\\
}

\maketitle
\thispagestyle{empty}

\newcommand{\eg}{\emph{e.g. }}
\newcommand{\ie}{\emph{i.e. }}

\begin{abstract}
This work presents an automatic human gender and age group recognition system based on human facial images. It makes an extensive experiment with row pixel intensity valued features and Discrete Cosine Transform (DCT) coefficient features with Principal Component Analysis and k-Nearest Neighbor classification to identify the best recognition approach. The final results show approaches using DCT coefficient outperform their counter parts resulting in a 99\% correct gender recognition rate and 68\% correct age group recognition rate (considering four distinct age groups) in unseen test images. Detailed experimental settings and obtained results are clearly presented and explained in this report.
\end{abstract}

\vspace{-0.5cm}
\Section{Introduction}
\vspace{-0.25cm}

Recently, automatic human gender and age recognition has started catching the attention of researchers due to its possible wide application pool. To name a few it can be used:
\begin{itemize}
\item in Human-Computer Interaction (HCI) systems to tune the context appropriately to suit the target person's gender and age.
\item to monitor specific gender or age restricted areas in surveillance systems
\item to make targeted advertising where the relevant advertisement/information to the audience can be channelled from electronic billboard systems.
\item in  automated biometric data acquisition and 
\item for content based search in which identifying the gender and age reduces the search space significantly.
\end{itemize}

Unfortunately non-intrusive recognition methods are very challenging due to the huge amount of ethnic and race variation within a target class. One must seek an approach that is capable of generalizing over a class while at the same time maintaining inter-class discrimination ability. Thanks to recent advances in the domain of computer vision, very sophisticated and advanced algorithms that utilize the rich image information from cameras to solve this problem are being reported. Even though automatic gender recognition can be performed using still images or videos capturing full or part of a person's body, most researchers have focused on facial image analysis \cite{one} due to ease and discriminative power.

In any recognition/classification problem involving images, two important considerations are the type of features extracted from the image and the kind of classifier used (e.g linear or non-linear) to discriminate between the different classes based on the extracted features.  The natural tendency is to use raw intensity pixel values as features, but these values do not capture the relationship between neighboring pixel values and hence by themselves have less discriminative information. To increase the discriminative capability features that capture the pattern within neighboring pixels must be extracted and used. Examples of these include statistical measures, local image histograms, spatial frequency coefficients, gradient histograms, and rectangular region differences.  

\begin{figure}[!h]
  \vspace{-0.2cm}
  \centering
  \includegraphics[width=0.45\textwidth]{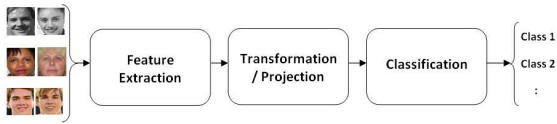}
  \vspace{-0.35cm}
  \caption{\footnotesize General facial image based recognition/classification approach.}
  \label{fig:syshwarch}
 \vspace{-0.2cm}
\end{figure}

On the other hand, the employed classifier must be capable of drawing a decision boundary between the different classes. It must not be sensitive to noise in the data and be able to generalize so as to perform well in unseen patterns. In the literature some of the classification techniques used in image based recognition problems include k-Nearest Neighbor (k-NN), AdaBoost, linear and non-linear Support Vector Machines (SVM), and Neural Networks. Often, to either increase the separability of the different classes or increase the representation power, it might be necessary to transform or project the extracted features into a new representation space before classification. Principal Component Analysis (PCA), Linear Discriminant Analysis (LDA), Clustering techniques are a few ways to achieve this. A general approach to recognition problem based on facial images is shown in Figure \ref{fig:syshwarch}. The exact choice of techniques in each stage depends on the specific problem and output requirements.

In facial image based gender recognition, the problem is to recognize whether a given facial image is male or female resulting in a two class recognition/classification problem. Various approaches have been reported in the literature which includes a shared rectangular feature with AdaBoost classifier \cite{two}, Discrete Cosine Transform (DCT) coefficients with PCA and Nearest Neighbor classifier \cite{three}. Makinen et. al. \cite{four} presented a detailed survey and evaluation of approaches that use SVMs, AdaBoost, and Artificial Neural Networks as classifiers with varying facial image features. In this work, two gender recognition systems that use raw intensity values and DCT coefficients as features with PCA and k-NN as classifier are implemented and experimental results reported. 

Another facial image based recognition considered in this work is age classification. Age classification can be seen as either a regression or a classification problem. If the problem is to estimate the age of a target person, then it is a regression problem. This approach requires a lot of training data with precise age information \cite{five} and is not privileged in this work. On the other hand, the age classification problem aims at identifying the age group of a target person \cite{six}. A similar framework shown in figure 1 can be used for age group recognition. In this work, age group classification is considered with four age groups corresponding to young age (15-20 years), adult age (21-40 years), middle age (41-60 years), and old age (above 60 years).

The rest of this paper is structured as follows: all necessary theoretical foundation for this work is presented in section 2, and then section 3 and section 4 present the methods, experiments, results and discussions corresponding to the implemented gender recognition and age group recognition systems respectively. Finally, the paper finalizes with a conclusion and further recommendations in section 5.

\vspace{-0.25cm}
\Section{Theoretical Background} \label{sec:framework}
\vspace{-0.25cm}

\subsection{Principal Component Analysis (PCA)}
Principal Component Analysis (PCA) is mathematically defined as an orthogonal linear transformation that transforms the data to a new coordinate system such that the greatest variance by any projection of the data comes to lie on the first coordinate (called the first principal component), the second greatest variance on the second coordinate, and so on \cite{eight}. Hence, it can be interpreted as a technique to reduce a large dimensional data to a smaller intrinsic dimensional feature space which are needed to describe the data economically. Because of this it has been used in image recognition and compression extensively. 
When using PCA for any image based recognition task, the following sequence of steps need to be taken.
\begin{itemize}	
\item Express features extracted from 2D images as 1D vector. Given M images each with N features $(f_1,...,f_N)$, this vector can be expressed as follows:
\begin{align}	\label{eqn:adddisn}
		x_i=[f_1,f_2,...,f_N ]^T; i=1,...,M
\end{align}
\item Mean center each vector by subtracting the total mean vector. 
\begin{align}	\label{eqn:adddisn}
	\tilde{x}_i=x_i-\bar{x}  ; \:\:\bar{x} = \frac{1}{M}\sum_{i=1}^{M}x_i
\end{align}
\item	Concatenate each mean centered vector column wise to construct a matrix of the data, W.
\begin{align}	\label{eqn:adddisn}
	W=[\tilde{x}_1  \tilde{x}_2 . . . \tilde{x}_M]
\end{align}
\item	Compute the covariance matrix of the data, $C=WW^T$, then the principal components are given by the Eigen vectors of the covariance matrix and their importance by the corresponding Eigen values.
\end{itemize}
Unfortunately, calculating the Eigen vectors of C which is an $NxN$ matrix can be computationally very expensive. A work around is to calculate the Eigen vectors and values of $\tilde{C}=W^TW$, and then pre-multiplying these vectors with $W$ to get the Eigen vectors of $\tilde{C}$. Since the dimension of $\tilde{C}$ is $MxM$, which is the size of total number of samples and usually $M<N$, then the calculation is less computationally demanding. In recognition problems, once the Eigen vectors are determined, every sample that needs to be recognized is projected onto this new representation space and compared with nearby sample classes. Due to its ability to extract principal data components, it is used in many applications including recognition, compression, and data mining \cite{nine}. 

\subsection{K-Nearest Neighbor (k-NN) Classification}
\begin{figure}[!h]
  \vspace{-0.2cm}
  \centering
  \includegraphics[width=0.28\textwidth]{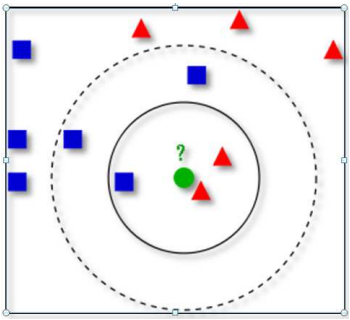}
  \vspace{-0.35cm}
  \caption{\footnotesize Example of k-NN classification. Example of k-NN classification. The test sample (green circle) should be classified either to the first class of blue squares or to the second class of red triangles. If k = 3 it is assigned to the second class because there are 2 triangles and only 1 square inside the inner circle. If k = 5 it is assigned to the first class (3 squares vs. 2 triangles inside the outer circle) (taken from \cite{seven})}
  \label{fig:two}
 \vspace{-0.2cm}
\end{figure}

K-Nearest Neighbor (k-NN) classification is a quite straightforward classification on which examples are classified based on the class of their nearest neighbors. The ‘k’ signifies the number of neighbors taken into account when making the classification decision. For a given instance, the algorithm checks the k nearest neighbors based on a specified distance (or similarity) measure, and the instance is classified as an object of the class with the maximum number of similarities. Since the training examples are needed at run-time, i.e. they need to be in memory at run-time, it is sometimes also called Memory-Based Classiﬁcation. Figure \ref{fig:two} taken from \cite{seven} visually illustrates a simplified k-NN classification example. The main advantage of k-NN is that it has a transparent easy to implement process. Though its sensitivity to redundant features, the need to store and use all training data makes it less appealing for real time applications.

\subsection{Image Features}
Image features are data extracted from an image that is used to represent the image in recognition/classification problems.  These features can be features that take each individual image pixels independently or features extracted in such a way to capture patterns represented in neighboring pixels. In our case extracting these features from the entire original image will help for the computation of the Eigen space. In this work, we have considered two features: row pixel intensity values and Discrete Cosine Transform (DCT) coefficients.
\paragraph{Pixel Intensity Value}
In this approach all image pixel data are taken as image feature by arranging them in a row or column wise manner. This extraction method will not guarantee that the attribute of all the extracted feature patterns will be the most relevant for classiﬁcation tasks. Compared to the other feature extraction methods this method is inefficient or it will not produce the best recognition rate for the classification task.

\paragraph{Discrete Cosine Transform (DCT) Coefficients}
Discrete cosine transform is used in several image processing applications including face recognition. We can use it for the extraction of important features which could be used for gender as well as age classification. DCT represents the given data in terms of cosine functions oscillating at different frequencies. It is capable of isolating gradually changing data (low frequency) from frequently changing data (high frequency).  From a 2D image of size $N_1xN_2$ , without normalization constants DCT coefficients $X_{(k_1,k_2)}$ are computed as follows:
\begin{align}	\label{eqn:adddisn2}
X_{(k_1,k_2)} = &\sum_{i=0}^{N_1-1}(\sum_{j=0}^{N_2-1}x_{i,j}*cos⁡[\frac{\pi}{N_2}(j+\frac{1}{2})k_2])\\ \nonumber
					 & *cos⁡[\frac{\pi}{N_1}(i+\frac{1}{2})k_1 ]
\end{align}

\begin{figure}[!h]
  \vspace{-0.2cm}
  \centering
  \includegraphics[width=0.3\textwidth]{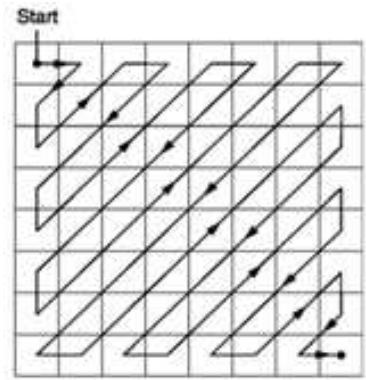}
  \vspace{-0.35cm}
  \caption{\footnotesize Illustration of zig-zag sampling of 2D DCT coefficients.}
  \label{fig:zigzag}
 \vspace{-0.2cm}
\end{figure}

After applying DCT in the face images we can see that most of the energy of the coefficients are concentrated in the few low-frequency components of the DCT or near the origin. Due to this fact selection of the important coefficients for the construction of feature vectors become an easy task. Selecting the coefficients by zigzag manner (as shown in figure \ref{fig:zigzag}) is the most used method.

\begin{table*}[!ht]%
\caption {\small Recognition rates of the Gender classification system using raw pixels as feature.}
\vspace{-0.2cm}
\label {table_1} 
\centering %
\scriptsize
\begin{tabular}{ ccccccc }
\toprule %
& 1-NN & 3-NN & 5-NN & 7-NN & 9-NN & Cluster Centroid\\ \midrule
Recognition Rate & 0.87 & 0.92 & 0.95 & 0.92 & 0.91 & 0.96\\\bottomrule
\end{tabular} 
\end{table*}

\vspace{-0.25cm}
\Section{Gender Classification} \label{sec:detection}
\vspace{-0.25cm}
\subsection{Overview}
\vspace{-0.25cm}
In facial image based gender recognition, the problem is to recognize whether a given facial image is male or female resulting in a two class recognition/classification problem. In this project, two techniques are implemented and their difference lies on the features used to develop the system. Referring to figure 1, the first approach uses raw pixels as extracted features, PCA as a projection, and k-NN classifier as a classification rule. The second approach uses DCT coefficients sampled in zigzag pattern as extracted features, PCA as a projection method, and k-NN classifier as a classification rule. In k-NN based classification, the Euclidean distance is used as a nearness measure. To train the system, a dataset of male and female images acquired from \cite{ten} and cropped and labelled to get the face region only, with an equal size of 128x128, is used. Sample original and cropped images taken from this dataset are shown in figure \ref{fig:gendersample} below.

\begin{figure}[!h]
  \vspace{-0.2cm}
  \centering
  \includegraphics[width=0.45\textwidth]{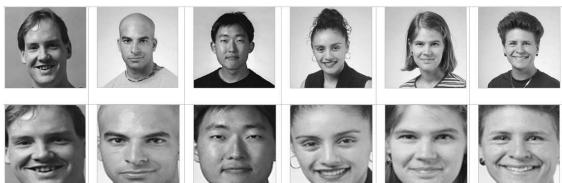}
  \vspace{-0.35cm}
  \caption{\footnotesize Sample images used for Gender Recognition.}
  \label{fig:gendersample}
 \vspace{-0.2cm}
\end{figure}

\subsection{Experiments}
Various experiments were carried out to determine which method yields the best recognition rate. In each experiment a training size of 100 male and 100 female images are used. To test the performance of the system, a test set consisting of 50 male and 50 female images that were not used in the training stage, is used. To quantize the recognition performance, recognition rate is calculated in each case.

\scriptsize
\begin{align}	\label{eqn:adddisn}
		Recognition \: Rate = \frac{correctly \: recognized \: test \: images}{total \: number \: of \: tested \: images}
\end{align}

\normalsize
In the first system which uses raw pixel features, experiments are carried out by varying the ‘k’ in the k-NN classifier from one up to nine in increments of two.  A slightly different technique which instead of taking the nearest neighbors to each instance, calculated the centroid of the Eigen faces corresponding to the male and female clusters separately and assigned a classification label corresponding to that of the nearest cluster centroid to each test instance is also experimented. Results to this variant case are reported with the label “Cluster Centroid”. In each case a sample feature vector has a dimension of 16384 corresponding to an image’s size of 128x128. In the system that uses DCT as features, the number of DCT coefficients is varied from 10 to 200, and each case, a set of tests with varying ‘k’, of the k-NN classifier, from one to nine in increments of two is performed. Similar to the raw pixel feature approach, an experimental test based on Cluster centroids is also performed. All pertinent results corresponding to each test are reported in section 3.3 below.

\subsection{Results and Discussions}
Recognition rates of the gender recognition system that uses raw pixels as features are shown in table 1. The best result is obtained with the cluster centroid approach which computes the cluster centroid of the Eigen faces in the male and female training and uses it to classify a test instance by assigning a class label corresponding to the nearest centroid.  The confusion matrix corresponding to this approach is shown in table 2.

Figure \ref{fig:dct_graph_1} shows plots of the gender recognition rates obtained using varying number of DCT coefficients and varying ‘k’ values of the k-NN classifier. It also shows the plot obtained using cluster centroid. The best result of 0.99 (99\% correct recognition rate) is obtained using a 7 nearest neighbor classifier with the first 134 DCT coefficients and using a 5 nearest neighbor classifier with the first 133 DCT coefficients. The resulting confusion matrix when using 5-NN classifier with 133 DCT coefficients is shown in table 3.

\begin{figure}[!ht]
  \vspace{-0.2cm}
  \centering
  \includegraphics[width=0.45\textwidth]{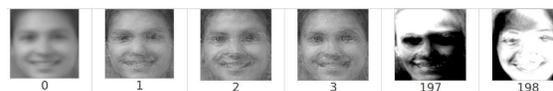}
  \vspace{-0.35cm}
  \caption{\footnotesize $0^{th}$, $1^{st}$, $2^{nd}$, $3^{rd}$, $197^{th}$, and $198^{th}$ Eigen faces obtained when using raw pixel features.}
  \label{fig:eigenface1}
 \vspace{-0.2cm}
\end{figure}

\begin{figure*}[!ht]
  \vspace{-0.2cm}
  \centering
  \includegraphics[width=0.8\textwidth]{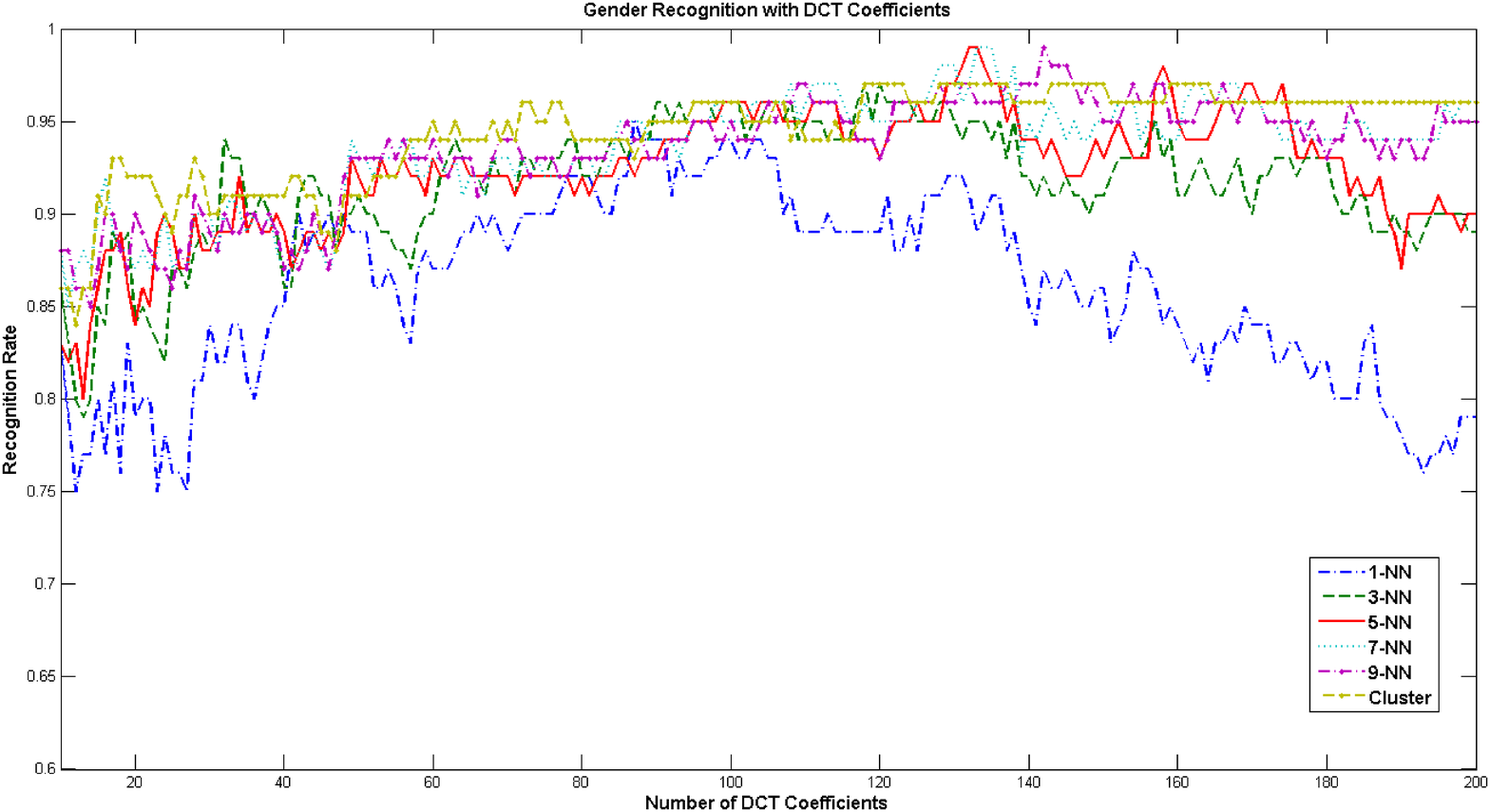}
  \vspace{-0.35cm}
  \caption{\footnotesize Gender recognition rate plot with varying number of DCT coefficients.}
  \label{fig:dct_graph_1}
 \vspace{-0.2cm}
\end{figure*}

\begin{figure}[!h]
  \centering
  \includegraphics[width=0.45\textwidth]{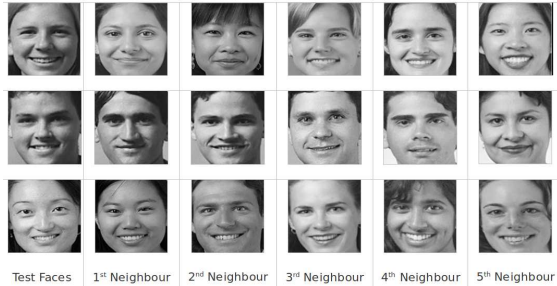}
  \vspace{-0.35cm}
  \caption{\footnotesize Exemplar correctly classified faces along with their 5 nearest neighbors. Obtained using 133 DCT coefficients with 5-NN classifier.}
  \label{fig:gendersampleresult}
 \vspace{-0.4cm}
\end{figure}

\begin{figure}[!h]
  \vspace{-0.2cm}
  \centering
  \includegraphics[width=0.45\textwidth]{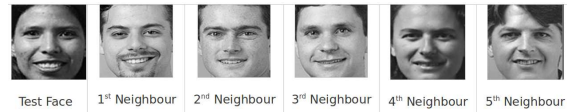}
  \vspace{-0.35cm}
  \caption{\footnotesize Wrongly classified female face and its 5 nearest neighbors.}
  \label{fig:gendersampleresult_w}
 \vspace{-0.2cm}
\end{figure}

Figure \ref{fig:eigenface1} shows the first 4 main and 2 last Eigen faces obtained when using raw pixel values as features.  These images are obtained by reshaping the Eigen faces into a two dimensional matrix with the dataset mean added on them. Exemplar correct classification samples obtained with the best recognition approach (DCT features with 5-NN classifier) are shown in figure \ref{fig:gendersampleresult}.  In each row the test face and the corresponding 5 neighbors are shown. The only wrongly classified test face corresponding to a female facial image is shown in figure \ref{fig:gendersampleresult_w} along with its 5 nearest neighbors.

\begin{table}[ht]
\begin{minipage}[b]{0.46\linewidth}\centering
\caption {\small Gender recognition confusion matrix with raw pixel features (cluster centroid).}
\scriptsize
\begin{tabular}{ ccc }
\toprule %
& Male & Female \\\midrule
Male & 49 & 1 \\\midrule
Female & 3 & 47\\\bottomrule
\end{tabular}
\end{minipage}
\hspace{0.1cm}
\begin{minipage}[b]{0.46\linewidth}
\centering
\caption {\small Gender recognition confusion matrix obtained using 133 DCT features with 5-NN classifier.}
\scriptsize
\begin{tabular}{ ccc }
\toprule %
& Male & Female \\\midrule
Male & 50 & 0 \\\midrule
Female & 1 & 49\\\bottomrule
\end{tabular} 
\end{minipage}
\end{table}

In this section, different experiments carried out to build an automatic gender recognition system have been presented.  The best recognition performance is obtained when using the first 133 DCT coefficients sampled in a zigzag manner and a 5-NN classifier. The system recognizes all but one test case correctly resulting in a 99\% correct recognition rate which is a huge improvement on random guessing with a 50\% chance. In general, experiments using DCT coefficients showed superior recognition performance than those using raw pixel values. This is expected as DCT features do not consider each pixel independently rather extract features that characterize the spatial intensity distribution amongst neighboring pixels thus by being able to distinctively represent general patterns.

\vspace{-0.25cm}
\Section{Age Group Classification} \label{sec:tracking}
\vspace{-0.25cm}
\subsection{Overview}
The second investigation of this work is age group classification. In this work, rather than trying to estimate the age of a person (a regression problem), age group recognition/classification is considered. This approach makes it feasible to use the same framework used for gender recognition with minor modifications. The first modification is the number of classes (classification categories). The following four age group categories are used.

\begin{figure}[!h]
  \vspace{-0.2cm}
  \centering
  \includegraphics[width=0.45\textwidth]{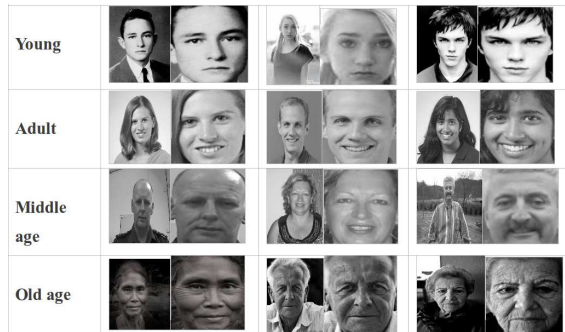}
  \vspace{-0.35cm}
  \caption{\footnotesize Sample original and cropped images from the age group training dataset.}
  \label{fig:gender_dataset_eg}
 \vspace{-0.2cm}
\end{figure}

\begin{table*}[!ht]%
\caption {\small Age group recognition rates obtained using raw pixel as features.}
\vspace{-0.2cm}
\label {table_1} 
\centering %
\scriptsize
\begin{tabular}{ ccccccc }
\toprule %
& 1-NN & 3-NN & 5-NN & 7-NN & 9-NN & Cluster Centroid\\ \midrule
Recognition Rate & 0.63 & 0.66 & 0.63 & 0.65 & 0.64 & 0.48\\ \bottomrule
\end{tabular} 
\end{table*}

\begin{enumerate}
\item Young age group : 15 - 20 years old
\item Adult age group : 21 - 40 years old
\item Middle age group :  41 - 60 years, and 
\item Old age: above 60 years of age.
\end{enumerate}
A training dataset consisting of 100 images in each group, totalling 400 images, evenly distributed between males and females is used. The images in the dataset are collected manually from the Internet; mostly from free online dating sites that included age in the person’s profile. The images are cropped to retain the face region only and labelled. Sample original and cropped images from this dataset are shown in figure \ref{fig:gender_dataset_eg}. 

\subsection{Experiments}
A similar experiment to that of gender classification is carried out for age group classification. The two approaches that use raw pixel data and DCT coefficients as features are tested considering ‘k’ nearest neighbors and nearness to cluster centroid classification techniques. Again in all NN classification, the Euclidean distance is used as a nearness measure. As stated before a total of 400 training images, 100 images in each category are used. For testing purposes, a total of 200 images, 50 images per category evenly distributed between males and females are used. Recognition rate is again used to quantify the performance of the each approach. When using DCT coefficients as features, experiments are carried out by varying the number of coefficients from 10 to 200 while the ‘k’ of the k-NN classifier is varied from 1 to 9 in steps of two units.  In the raw pixel case, each feature vector has a dimension of 16384 resulting from the 128x128 facial images. 

\subsection{Results and Discussions}
Results obtained when using raw pixel values as features are shown in table 4.  The best result in this case obtained when using 3-NN as a classification rule and it can categorize unseen faces with 66\% accuracy. The confusion matrix corresponding to this best raw pixel as features based classifier is shown in table 5. As can be seen from this table, most of the errors occur in the young and middle age categories.  This system confuses young with adult and middle age with young severely.

When using DCT coefficients as features, it can be seen from figure \ref{fig:dct_age_group} that the best recognition performance of 68\% is obtained when considering the first 91 DCT coefficients and 7 nearest neighbors (7-NN classification).  The confusion matrix corresponding to this best classifier is shown in table 6. Compared to the best classifier using raw pixel values, this classifier improves upon the adult and middle age, age groups but does worse with	the young and old age, age groups. It confuses young with adult and old with middle age severely. 


\begin{table}[ht]
\caption {\small Age group recognition confusion matrix using raw pixel features with3-NN classifier.}
\scriptsize
\begin{tabular}{ ccccc }
\toprule %
	& Young & Adult & Middle Age & Old Age \\\midrule
Young & 28 & 9 & 7 & 6 \\\midrule
Adult & 7  & 38 & 2 & 3 \\\midrule
Middle Age & 10 & 5 & 28 & 7 \\\midrule
Old Age & 3 & 4 & 5 & 38 \\\bottomrule
\end{tabular}
\end{table}

\begin{table}
\centering
\caption {\small Age group recognition confusion matrix using 91 DCT coefficients and 7-NN classifier.}
\scriptsize
\begin{tabular}{ ccccc }
\toprule %
	& Young & Adult & Middle Age & Old Age \\\midrule
Young & 26 & 16 & 2 & 6 \\\midrule
Adult & 2  & 44 & 3 & 1 \\\midrule
Middle Age & 3 & 6 & 34 & 7 \\\midrule
Old Age & 1 & 1 & 16 & 32 \\\bottomrule
\end{tabular} 
\end{table}

\begin{figure}[!h]
  \vspace{-0.2cm}
  \centering
  \includegraphics[width=0.45\textwidth]{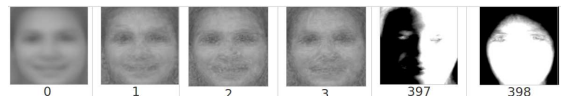}
  \vspace{-0.35cm}
  \caption{\footnotesize $0^{th}$, $1^{st}$, $2^{nd}$, $3^{rd}$, $397^{th}$, and $398^{th}$  Eigen faces obtained when using raw pixel features for age group classification.}
  \label{fig:age_group_eigenface}
 \vspace{-0.2cm}
\end{figure}

\begin{figure*}[!ht]
  \vspace{-0.2cm}
  \centering
  \includegraphics[width=0.8\textwidth]{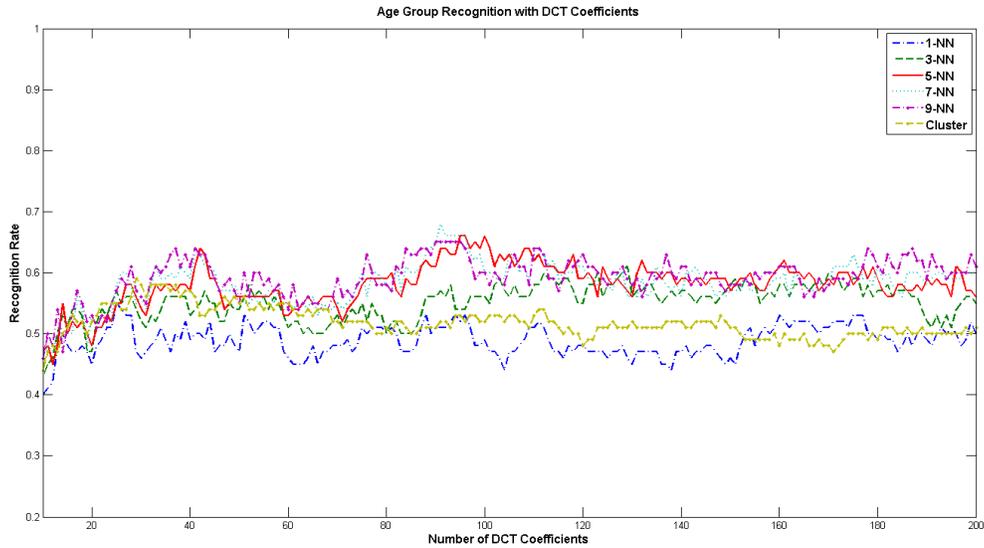}
  \vspace{-0.35cm}
  \caption{\footnotesize Age group recognition rate plot with varying number of DCT coefficients and k-NN classifiers.}
  \label{fig:dct_age_group}
 \vspace{-0.2cm}
\end{figure*}

Figure \ref{fig:age_group_eigenface} shows the first four and the last two Eigen faces obtained when considering raw pixel values. Each image is generated by reshaping the Eigen vectors in a 2D matrix and adding the subtracted image mean values.

\begin{figure}[!h]
 \vspace{-0.2cm}
  \centering
  \includegraphics[width=0.45\textwidth]{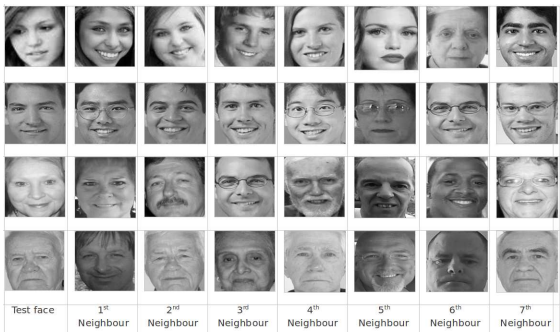}
  \vspace{-0.35cm}
  \caption{\footnotesize Examples of correctly classified age groups.}
  \label{fig:agegroup_correct_classification}
 \vspace{-0.2cm}
\end{figure}

\begin{figure}[!h]
  \centering
  \includegraphics[width=0.45\textwidth]{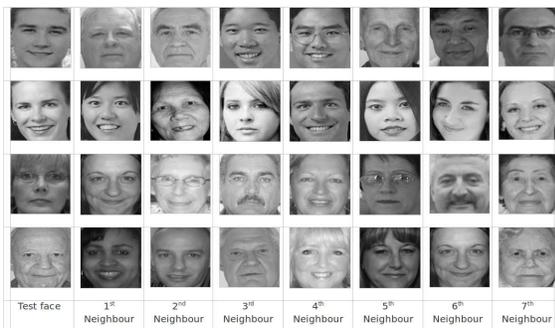}
  \vspace{-0.35cm}
  \caption{\footnotesize Sample examples taken from wrongly classified age groups.}
  \label{fig:agegroup_wrong_classification}
 \vspace{-0.2cm}
\end{figure}

Sample correct and incorrect age group classifications corresponding to the best classifier, i.e. with 91 DCT coefficients and 7-NN classifiers, is shown in figure \ref{fig:agegroup_correct_classification} and \ref{fig:agegroup_wrong_classification} consecutively. Each row corresponds to a test face and its 7 nearest neighbors.  The exemplar cases are taken from each category with the 1st row corresponding to young, 2nd to that of adult, 3rd to that of middle age, and 4th to that of old age.

Similar to the gender recognition task, the approach which uses image DCT coefficients as features outperforms the approach which uses raw intensity pixels as features even if the improvement is not very high.  Knowing there are of four classes in the presented age group recognition work and a random guessing will have a 25\% accuracy rate, the 68\% recognition rate accomplished using the first 91 DCT coefficients and a 7-NN classifier is very high and promising.

\begin{figure}[!h]
  \vspace{-0.2cm}
  \centering
  \includegraphics[width=0.45\textwidth]{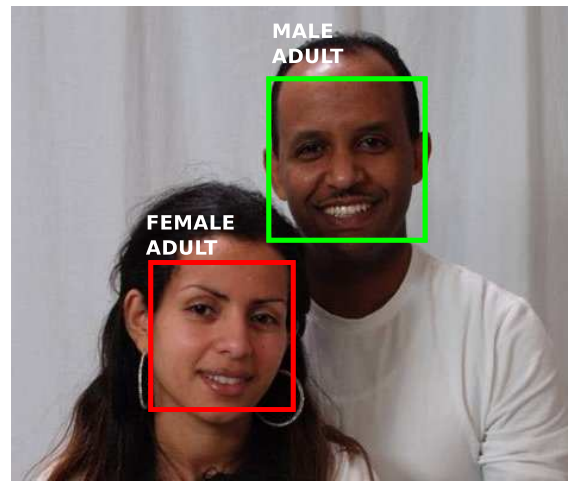}
  \vspace{-0.35cm}
  \caption{\footnotesize Real time implementation sample test output.}
  \label{fig:real_time}
 \vspace{-0.2cm}
\end{figure}

\normalsize
\vspace{-0.3cm}
\Section{Real Time Implementation} \label{sec:conclusion}
\vspace{-0.3cm}
A real time application that employes the best identified classification approaches, namely 133 DCT coefficients with 5-NN classifier for gender
recognition and DCT coefficients and 7-NN classifier for age group classification, is implemented in c using the open-source {OpenCv} \cite{opencv_library} library.
The final system executes moderately at 10 fps on an Intel core 2 Duo computer. Faces are automatically detected in the image using OpenCv's implementation
of the Viola and Jones \cite{viola} real time face detector. The automatically detected faces are then used as input into our implemented system to determine
their gender and age group. A sample successful output is shown in figure \ref{fig:real_time} with an image taken from the Internet.

\normalsize
\vspace{-0.3cm}
\Section{Conclusions and Recommendations} \label{sec:conclusion}
\vspace{-0.3cm}
In this work two recognition tasks, namely: gender recognition and age group recognition, based on facial image have been implemented and experimented upon. Experiments were carried out using two types of features, raw pixel data and DCT coefficients, PCA and k-NN classifier. The final results show that in both tasks, approaches which use image DCT coefficients as features perform best.  In gender recognition the best result of 99\% recognition rate is accomplished using a 133 DCT features with 5-NN classifier and a 68\% top recognition rate is obtained in age group recognition using 91 DCT features and 7-NN classifier.
Possible further improvements that ought to be considered to improve the performance of the age group recognition system beyond its current state include increasing the dataset to help the system generalize better, and use sophisticated classification method like Support Vector Machines (SVMs) and/or AdaBoost by mixing the currently considered and if possible additional features.

\vspace{-0.3cm}
\footnotesize
\nocite{ex1,ex2}
\bibliographystyle{latex12}
\bibliography{bibliography_mod.bib}
\end{document}